\documentclass[10pt,twocolumn,letterpaper]{article}

\usepackage[accsupp]{axessibility}  % Improves PDF readability for those with disabilities.
\usepackage[pagenumbers]{wacv} % To force page numbers, e.g. for an arXiv version

\usepackage{graphicx}
\usepackage{amsmath}
\usepackage{amssymb}
\usepackage{booktabs}
\usepackage{booktabs, multirow} % for borders and merged ranges
\usepackage{soul}% for underlines
\usepackage[table]{xcolor} % for cell colors
\usepackage{changepage,threeparttable} % for wide tables
\usepackage{tabularx}
\newcolumntype{Y}{>{\centering\arraybackslash}X}
\usepackage{xcolor}

\usepackage[pagebackref,breaklinks,colorlinks]{hyperref}

\usepackage[capitalize]{cleveref}
\crefname{section}{Sec.}{Secs.}
\Crefname{section}{Section}{Sections}
\Crefname{table}{Table}{Tables}
\crefname{table}{Tab.}{Tabs.}

\def\wacvPaperID{312} % *** Enter the WACV Paper ID here
\def\confName{WACV}
\def\confYear{2024}

\begin{document}

%%%%%%%%% TITLE
\title{Learning 
Class and Domain Augmentations for Single-Source Open-Domain Generalization}

% \author{Prathmesh Bele\\
% Indian Institute of Technology Bombay\\
% Mumbai, India - 400076\\
% {\tt\small prathmeshbele.iitb@gmail.com}
% \and
% Avigyan Bhattacharya\\
% Jadavpur University\\
% Kolkata, India - 700032\\
% {\tt\small avigyanbhattacharya123@gmail.com}
% }

\author{Prathmesh Bele$^{1}$ \quad \hspace{-0.45cm} Valay Bundele$^{1}\thanks{equal contribution}$ \quad \hspace{-0.4cm} Avigyan Bhattacharya$^{1*}$ \quad \hspace{-0.45cm} Ankit Jha$^{1}$ \quad \hspace{-0.45cm} Gemma Roig$^{2}$ \quad \hspace{-0.45cm} Biplab Banerjee$^{1}$ \vspace{0cm} \\
{\normalsize $^1$Indian Institute of Technology Bombay, India} \quad
{\normalsize $^2$Goethe University Frankfurt, Germany} \quad \\
{\tt \small {prathmeshbele.iitb@gmail.com, valaybundele@gmail.com, avigyanbhattacharya123@gmail.com}}\\
{\tt \small{ankitjha16@gmail.com, roig@cs.uni-frankfurt.de, getbiplab@gmail.com}}
}

\maketitle
\thispagestyle{empty}

%%%%%%%%% ABSTRACT
\begin{abstract}
Single-source open-domain generalization (SS-ODG) addresses the challenge of labeled source domains with supervision during training and unlabeled novel target domains during testing. The target domain includes both known classes from the source domain and samples from previously unseen classes. Existing techniques for SS-ODG primarily focus on calibrating source-domain classifiers to identify open samples in the target domain. However, these methods struggle with visually fine-grained open-closed data, often misclassifying open samples as closed-set classes. Moreover, relying solely on a single source domain restricts the model's ability to generalize.
To overcome these limitations, we propose a novel framework called \textsc{SODG-Net} that simultaneously synthesizes novel domains and generates pseudo-open samples using a learning-based objective, in contrast to the ad-hoc mixing strategies commonly found in the literature. Our approach enhances generalization by diversifying the styles of known class samples using a novel metric criterion and generates diverse pseudo-open samples to train a unified and confident multi-class classifier capable of handling both open and closed-set data.
Extensive experimental evaluations conducted on multiple benchmarks consistently demonstrate the superior performance of \textsc{SODG-Net} compared to the literature.

\end{abstract}
\vspace*{-6mm}
%%%%%%%%% BODY TEXT
\section{Introduction}
\vspace*{-1mm}

Deep learning models often face significant performance degradation when confronted with domain shift \cite{1,2}, which occurs when the training and test data originate from different distributions. To address this issue, domain generalization (DG) has been proposed as a means to enable models to generalize to unknown target domains \cite{3,4}. In the classical DG setup, the assumption is made that multiple source domains are accessible during training, known as multi-source DG (Multi-DG). Notable advancements have been achieved in Multi-DG over the past decades, utilizing techniques such as domain alignment, meta-learning, domain augmentation, and self-supervision \cite{5,6,7,8,9,10}.

However, collecting and labeling data from multiple domains can be expensive, making Multi-DG less practical in real-world scenarios. In contrast, Single-Source Domain Generalization (Single-DG) is a more challenging and realistic setting that has been less explored. In Single-DG, only one source domain is available for training, making it difficult for the model to learn domain-invariant information due to the absence of domain comparisons. As a result, the model is susceptible to overfitting the domain-specific signals present in the single source domain. In recent years, some efforts \cite{11,18,19} have focused on domain augmentation for addressing Single-DG. \textit{However, it should be noted that accountable novel domain augmentation from a single source domain is a non-trivial problem.}
\begin{figure}
\begin{center}
\includegraphics[width=0.875\linewidth]{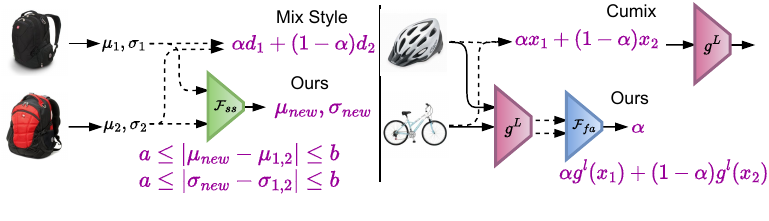}
\end{center}
\vspace{-0.8cm}
   \caption{Differences between existing domain and sample interpolation methods (Mixstyle \cite{zhou2021domain}, Cumix \cite{mancini2020towards} ) and our proposed \textsc{SODG-Net}  for style/sample synthesis. In both \cite{zhou2021domain} and \cite{mancini2020towards}, $\alpha$ is sampled from the dist. $\mathcal{B}(\Gamma)$, where we propose a learning based scheme. $\mathcal{F}_{ss}$ and $\mathcal{F}_{fa}$ denote the proposed style synthesis and feature aggregation blocks, respectively.}
   \vspace{-0.6cm}
\label{fig:teaser}
\end{figure}

While both Multi-DG and Single-DG are considered closed-set models, meaning that the source and target domains share the same set of categories, this assumption is often unrealistic in real-world applications. In practice, there is often a lack of prior information available for the target domain, leading to the possibility of encountering samples from previously unknown classes alongside the known class samples. This scenario has given rise to the problem of Open-Set Domain Generalization (ODG), which is applicable in both single and multi-DG settings.
Despite the significant importance of ODG in various applications such as self-driving cars, remote sensing, and medical imaging, there is a scarcity of ODG models in the existing literature \cite{20,12}. To this end, SS-ODG poses even greater challenges due to the scarcity of source domain styles and the presence of novel class samples in the target domain.
It is worth noting that \textsc{CrossMatch} \cite{12}, the only existing single-source ODG model, tackles this issue by employing an adversarial augmentation strategy to synthesize out-of-distribution samples as pseudo-unknown data from the source domain. Furthermore, a majority voting-based classification strategy is utilized to identify open samples during testing.

Although \textsc{CrossMatch} demonstrates superior performance when combined with existing Single-DG methods, it does have a few potential bottlenecks:
   \textbf{i)} The generation of pseudo-open samples aims for a significant separation from the known class data, resulting in a coarse relationship between the known and pseudo-open samples. This approach may hinder the classification of fine-grained target open samples that have a closer relationship with the source domain data.
 \textbf{ii)}   \textsc{CrossMatch} primarily focuses on the open-set classification task and lacks emphasis on improving the performance of closed-set classes. It does not support augmentation for the known-class samples and relies on existing Single-DG models for this purpose. 
  \textbf{iii)}  The multi-binary classifier used in \textsc{CrossMatch} neglects the class-level correlations, leading to performance degradations for fine-grained known and unknown samples.

The discussions highlight three significant research gaps in addressing SS-ODG:
\textbf{i)}  \textit{How can we effectively synthesize diverse novel domains for the known classes when we only have a single source domain?}
  \textbf{ii)}   \textit{How can we generate pseudo-open samples that can be used to train a unified classifier capable of handling both open and closed classes in a confident fashion?} and \textbf{iii)} \textit{Enhancing classification confidence within an ODG setting is a non-trivial task, especially considering the absence of prior knowledge about the target domain.}
    Existing style interpolation techniques \cite{zhou2021domain} are not suitable for \textbf{(i)} due to the constraint of having only one source domain in SS-ODG, and style extrapolation methods \cite{18,22} may be challenging and costly to apply. For \textbf{(ii)}, traditional mix-up strategies \cite{23, mancini2020towards} based on stochastic \textsc{Beta} sampling may not always generate open samples reliably.  \textit{Given these challenges, there is a clear need for learning-based approaches to address \textbf{(i)}-\textbf{(iii)}.}

\noindent \textbf{Our proposed \textsc{SODG-Net}:} In this paper, we present a comprehensive solution to address these gaps by introducing a novel end-to-end model called \textsc{SODG-Net}. Our approach involves synthesizing novel style primitives and generating representative open-set samples by separately leveraging domain and content knowledge extracted from the source domain images at the feature space, and further training a unified classifier for the closed and open classes while optimizing the confidence of its predictive ability.

We recognize that the feature response statistics obtained from different layers of a vision encoder capture valuable domain-dependent information \cite{instyle}. To capitalize on this, we propose a \textit{style generator} module within \textsc{SODG-Net}. This module takes style information from two distinct images and produces a new style vector that maintains a significant margin from the input data's style vectors (Fig. \ref{fig:teaser}). This approach diverges from the conventional style interpolation methods \cite{jin2021style, luo2020adversarial,zhou2021domain}. While those methods simply blend the feature statistics of two input domain images through a basic convex combination, they lack the guarantee of producing distinct styles. Likewise, techniques involving style extrapolation \cite{li2018learning} are restricted by the inherently uncertain nature of the problem, limiting their ability to generate a substantial variety of styles. Our solution to both these challenges lies in the innovative margin-based learning framework we introduce for style hallucination.

Furthermore, when generating the pseudo-open data, we depart from traditional ad-hoc mix-up techniques \cite{zhang2017mixup}. Instead, we introduce the concept of combining the feature embeddings of two distinct-class images using learnable feature weightings (Fig. \ref{fig:teaser}). We then impose a constraint on the generated embedding, ensuring it is classified as belonging to the unknown class. By doing so, we establish a precise pseudo-open-closed boundary, facilitating improved identification of potential unknown-class samples.
We identify our \textbf{major contributions} as follows,

\noindent -    We address the SS-ODG problem through a data augmentation perspective, introducing \textsc{SODG-Net}, complementing the literature. \textsc{SODG-Net} can effectively learn to synthesize novel domain primitives and representative pseudo-open-class embeddings simultaneously.

\noindent -    To ensure the distinctiveness of the generated domains both from the source domain and among themselves, we introduce a novel margin objective and a noisy-injected domain diversification strategy. Moreover, we put forth a weight learning approach that facilitates the fusion of feature embeddings from a pair of images, enabling the creation of a unified feature representation capturing the open space. Finally, we introduce intuitive objectives to increase the confidence of the predictions for both the closed and open classes, thus better handling visually alike samples.

\noindent -    Thorough experiments were conducted on four benchmark datasets to evaluate \textsc{SODG-Net}'s performance for SS-ODG. On average, we observed a significant improvement of around 1-14\% in the average h-score metric.

%------------------------------------------------------------------------
\vspace*{-2mm}

\section{Related Works}

\vspace*{-1mm}

\noindent{\textbf{Open-Set Domain Adaptation}}: OSDA is a related but different problem from ODG. OSDA considers a single labeled source domain and an unlabeled target domain, where the target domain contains additional samples from previously unknown classes. OSBP \cite{28} adopts an adversarial technique to train representations that effectively distinguish unknown target samples. STA \cite{liu2019separate} employs a binary classifier to finely separate all target samples and applies weight adjustment to mitigate the negative influence of unfamiliar target samples. Additionally, there exist other approaches such as TIM \cite{kundu2020towards}, SHOT \cite{liang2020we}, JPOT \cite{xu2020joint}, to name a few. In contrast, the SS-ODG problem assumes that source domain is only present during training, while unlabeled target domain appears only during inference. Hence, OSDA methods cannot be directly applied to SS-ODG.

\noindent{\textbf{Open Domain Generalization}}: On the other hand, ODG poses a more challenging and realistic problem, initially introduced by Shu et al. \cite{20}. In this problem, the source domains, which are multiple in number, and the singular target domain have different label spaces. The objective is to develop a model that can learn from multiple sources of data and utilize that knowledge to classify new data points into known classes from the source data or into new, unknown classes. To tackle this problem, \cite{20} proposed an approach that involves augmenting each source domain with missing class and domain knowledge through a novel \textsc{Dirichlet mixup} and distilled soft-labeling technique. Subsequently, a meta-learning technique was employed over the augmented domains to acquire open-domain generalizable representations. While \cite{20} is a multi-source ODG setup, \textsc{CrossMatch} \cite{12} addresses a similar yet more complex problem, where there is only one source domain available for training but multiple unseen target domains for testing. \cite{12} considered adversarial learning to hallucinate pseudo-open samples and deployed a multi-binary classifier, which is deemed to classify a given sample as belonging to a particular class or the open space. 

While \cite{12} serves as the closest existing framework to ours, there are significant differences between our approach and \cite{12}. These differences primarily lie in two aspects: \textbf{i)} we recognize the challenges associated with extrapolating new domains using the unstable adversarial approach employed in \cite{12}. Instead, we propose a margin-based objective that generates novel domain primitives, ensuring their distinguishability from the source domain, and \textbf{ii)} \cite{12} relies on a voting-based classifier, which is susceptible to misclassification due to a lack of confidence. In contrast, we address this issue by directly learning an open-set classifier and introducing a learning-driven method for synthesizing representative pseudo-open data embeddings and explicitly enhancing the confidence of the predictions through novel losses. As a result, we are better equipped to tackle the SS-ODG problem compared to \cite{12}.

\noindent{\textbf{Augmentation Techniques:}} Previous works on domain or data augmentations, such as Goodfellow et al. \cite{gan}, Kingma et al. \cite{varbayes}, and Zhang et al. \cite{zhang2017mixup}, have utilized variational autoencoders, GANs, or mixing strategies. Rahman et al. \cite{rahman2019multi} employed ComboGAN \cite{anoosheh2018combogan} to generate new samples and used various metrics to minimize the discrepancy between the generated and real data. In addition to generative models, mix-up techniques \cite{zhang2017mixup, mancini2020towards} have been used to create new samples by interpolating between pairs of samples and their labels using \textsc{Beta} or \textsc{Dirichlet} sampling. It has also been adopted in the areas of semi-supervised \cite{sohn2020fixmatch} and unsupervised learning \cite{xie2020unsupervised} combined with pseudo-labeling and consistency regularization, respectively. \cite{verma2019manifold} extends the mixup approach to the hidden states of deep neural networks while \cite{vu2019dada} applies it to the domain adaptation task in semantic segmentation. 

Another approach by \cite{22} involved training a GAN-based model using optimal transport to synthesize data from pseudo-novel domains. DLOW \cite{gong2019dlow} bridged different domains by generating a continuous sequence of intermediate domains. MixStyle \cite{zhou2021domain} randomly added batch-norm statistics of a pair of cross-domain samples.

Our approach to domain synthesis significantly departs from established domain interpolation techniques such as \cite{zhou2021domain,luo2020adversarial}. Firstly, we introduce a novel approach to learning distinct styles by employing a margin-based objective tailored specifically for the SS-ODG problem. In contrast, the applicability of \cite{zhou2021domain,luo2020adversarial} is more aligned with the Multi-DG problem and remains ad-hoc in nature.
Furthermore, our incorporation of a diversification criterion ensures the creation of conspicuously distinctive generated domains. Along similar lines, our feature mixing strategy, facilitated by a neural network, outperforms traditional mix-up methods \cite{mancini2020towards, zhang2017mixup}. This is due to our deliberate integration of learning-based criteria during the determination of mixing coefficients. The impact of these distinguishing features is evident in the results we present (Section \ref{sec:exp}).
%-------------------------------------------------------------------------

\vspace*{-2mm}
\section{Methodology}\label{sec:method}
\vspace*{-1mm}

\begin{figure*}
\begin{center}
\includegraphics[width=0.75\textwidth]{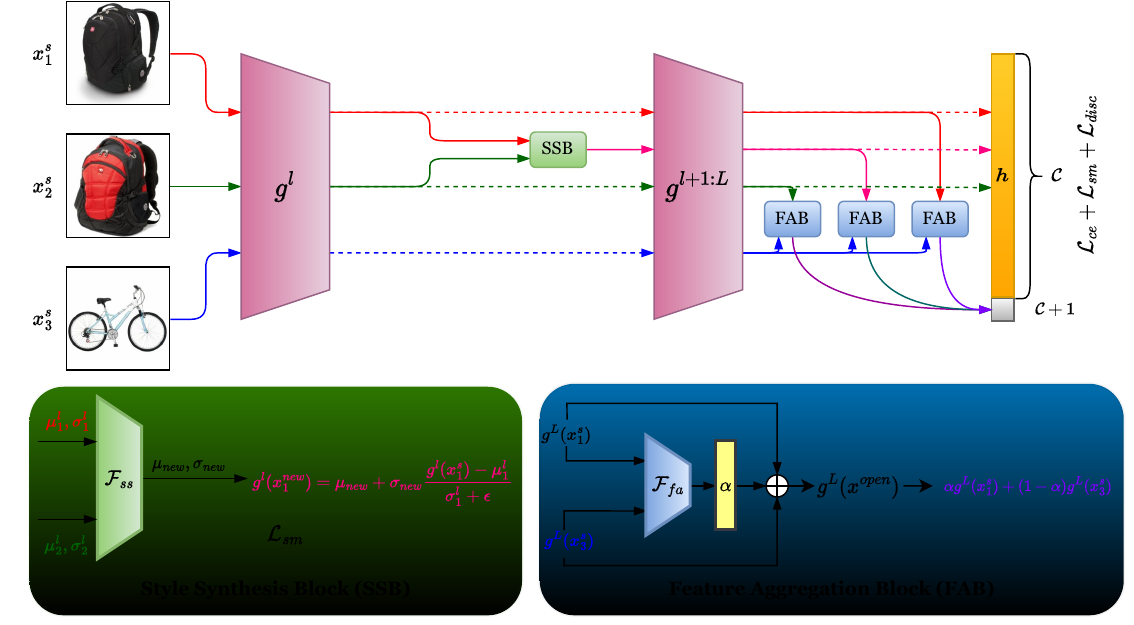}
\end{center}
\vspace{-0.5cm}
   \caption{\textbf{The model architecture of \textsc{SODG-Net}}. The model consists of a feature extractor $g$ and an open-closed classifier $h$. $l$ denotes an encoder layer of $l$ which consists of a total of $L$ layers. $\mathcal{F}_{ss}$ and $\mathcal{F}_{fa}$ denote the proposed style synthesis and feature aggregation blocks. The figure also depicts the loss functions, which are explained in Section \ref{sec:method}.}
\label{fig:SODG-NET}
\vspace{-0.5cm}
\end{figure*}

The SS-ODG problem revolves around a labeled source domain $\mathcal{S}$, which consists of training data $\mathcal{D}^s = \{x_i^s, y_i^s\}_{i=1}^{\mathcal{N}_s}$. Here, $x^s \in \mathcal{X}^s$ represents input images sampled from the domain-specific distribution $\mathcal{P}(\mathcal{X}^s)$, and $y^s \in \mathcal{Y}^s$ represents the corresponding label set for $\mathcal{S}$. During testing, the model encounters unlabeled samples from a previously unknown target domain $\mathcal{T}$: $\mathcal{D}^t = \{x_j^t\}_{j=1}^{\mathcal{N}_t}$. The label set $\mathcal{Y}^t$ encompasses $\mathcal{Y}^s$, meaning $\mathcal{Y}^s \subset \mathcal{Y}^t$. Our objective is to learn a parameterized prediction model $f =  g \circ h$, where $g$ represents the generic feature extractor and $h$ denotes the $\mathcal{C}+1$ class classifier. Here, $|\mathcal{Y}^s| = \mathcal{C}$, indicating the number of classes in $\mathcal{Y}^s$.

%\vspace*{-2mm}

\subsection{Overview of \textsc{SODG-Net}}

%\vspace*{-2mm}

In this section, we present the architecture and working principles of \textsc{SODG-Net}, as depicted in Fig. \ref{fig:SODG-NET}. Our main objectives during the training of \textsc{SODG-Net} can be summarized as follows:
\textbf{i)} Ensuring that the feature extractor $g$ is domain-generic, capable of extracting semantically coherent features from diverse visual domains.
\textbf{ii)} Enabling the classifier $h$ to simultaneously classify known-class samples from $\mathcal{T}$ into one of the $\mathcal{C}$ categories while consistently assigning a common label $\mathcal{C}+1$ to all samples corresponding to the classes in $\mathcal{Y}^t - \mathcal{Y}^s$. \textbf{iii)} To improve the confidence of the predictions of $h$.

To ensure the domain independence of $g$, we propose augmenting $\mathcal{D}^s$ with on-the-fly generated pseudo-domains and training $g$ on the augmented domain set. However, synthesizing new domains from a single source domain is a challenging task. To address this, we introduce novel style synthesis blocks $\mathcal{F}_{ss}$ within $g$. These blocks learn to generate new domain primitives by leveraging the domain properties of a pair of input images.
It is important to note that the instance-wise feature statistics, specifically the mean ($\mu$) and standard deviation ($\sigma$) from a CNN layer, capture domain artifacts, and different layers of $g$ capture styles at different abstractions. Hence, it is possible to plug in $\mathcal{F}_{ss}$ after the $l^{th}$ encoder layer of $g$. We delve into the details of the style synthesis block and its functioning in Section \ref{sec:ss}.

Additionally, we aim to generate feature embeddings for the pseudo-open samples towards training $h$. To achieve this, we introduce a feature aggregation and novel feature generation network, denoted as $\mathcal{F}_{fa}$. This network combines the feature embeddings of two images with different class labels, producing a potential unknown class feature. The workings of $\mathcal{F}_{fa}$ are elaborated further in Section \ref{sec:fa}.

%\vspace*{-1mm}

\subsection{Style synthesis block (SSB) to generate novel domains}\label{sec:ss}

\vspace*{-1mm}

The style synthesis block utilizes the style information from two input images to generate novel style characteristics that are significantly different from the input image styles.
For a given pair of images, $x_1^s$ and $x_2^s$, belonging to the same class, we calculate the mean $\mu$ and standard deviation $\sigma$ of their respective feature maps, $g^{l}(x_1^s)$ and $g^{l}(x_2^s)$, at the $l^{th}$ encoder layer. Let us denote these as $(\mu_i^l, \sigma_i^l)$, where $i \in {1,2}$.
To obtain $\mu_{new}, \sigma_{new}$, the mean and standard deviation vectors representing the synthesized style, we add noise sampled from a standard normal distribution $\mathbb{N}(0,1)$ to each of the $(\mu_i^l, \sigma_i^l)$ and concatenate the mean and standard deviation vectors of the input images and pass them through $\mathcal{F}_{ss}^l$, denoting the style synthesis module coupled with $g^l$. \textit{Our idea is to generate novel styles for each class separately, hence, we consider $(x_1^s, x_2^s)$ from the same class.}
The process can be expressed as follows:

\[ \mu_{new}, \sigma_{new} \leftarrow \mathcal{F}_{ss}^l([\mu_{1}^l + \delta_1;\sigma_{1}^l+ \delta_2;\mu_{2}^l+ \delta_3;\sigma_{2}^l+ \delta_4])\]

With the addition of the randomly sampled $\delta$ to the input vectors, we ensure generative modeling of the styles.
Furthermore, to guarantee the distinctiveness of the generated $[\mu_{new}, \sigma_{new}]$ from the input $[\mu_1, \sigma_1, \mu_2, \sigma_2]$, we introduce a margin loss criterion, denoted as $\mathcal{L}_{sm}$ (detailed in Eq. \ref{eq:Hingeeqn}). In this regard, we begin by applying instance normalization to the feature encoding $g^{l}(x_1^s)$. This step is crucial as it helps remove the original style characteristics present in the feature encoding of $x_1^s$ \cite{61,62,63}.
After the instance normalization step, we utilize the newly generated mean and standard deviation vectors, ${\mu}_{new}$ and ${\sigma}_{new}$, to modify the style of $g^l(x_1^s)$. This modification results in a sample with a novel style, distinct from the input images' styles but with identical semantic properties. $\epsilon$ is a small constant in Eq. \ref{eq:1} to ensure numerical stability.

\begin{equation}
\centering
g^l(x_1^{new}) = \mu_{new} + \sigma_{new} (\frac{g^{l}(x_{1}^s)-\mu_{1}}{\sigma_{1}+\epsilon})
\label{eq:1}
\end{equation}

%\vspace*{-2mm}

\subsection{Feature aggregation block (FAB) for open sample generation}\label{sec:fa}

\vspace*{-1mm}

The feature aggregation block, $\mathcal{F}_{fa}$, in contrast to $\mathcal{F}{ss}$, aims to synthesize samples representing out-of-distribution arbitrary classes by mixing the content features of samples from two different classes in $\mathcal{Y}^s$. Unlike mix-up based strategies \cite{mancini2020towards}, our approach learns the mixing coefficient $\alpha$ using $\mathcal{F}_{fa}$, with the intention of classifying the synthesized feature as belonging to the $\mathcal{C}+1$ category directly.
To achieve this, we utilize the output feature maps from the last convolutional block of $g$, denoted as $g^{L}(x_1^s)$ and $g^{L}(x_3^s)$, corresponding to images $x_1^s$ and $x_3^s$ with different class labels. By concatenating these feature maps and passing them through $\mathcal{F}_{fa}$, we obtain a weighting vector $\alpha$. This vector determines the contribution of each image's feature maps to generate a novel class sample, computed as $g^L(x^{open}) = \alpha \odot g^L(x_1^s) + (1- \alpha) \odot g^L(x_3^s)$. The mixing process, utilizing higher-level feature maps that capture abstract content information, allows the generation of samples representing novel classes by leveraging diverse input samples' content information.
Additionally, to enforce style variations in the generated open samples, we use Eq. \ref{eq:1} to update their style. This ensures that the synthesized samples exhibit distinct style characteristics, further enhancing the diversity and realism of the generated open samples.

%\vspace*{-1mm}

\subsection{Loss functions, training, and inference}\label{subsec: Loss}

\vspace*{-1mm}

A combination of multiple losses is employed to train both $g$ and $h$. The primary objective, considering the features of the closed and generated open samples, is the cross-entropy loss, which can be expressed as follows:
\begin{equation}\label{CEeqn}
 \mathcal{L}_{ce}  = \underset{\mathcal{P}(\mathcal{X}^s \cup \{x^{open}\}, \mathcal{Y}^s \cup \mathcal{C}+1)}{\mathbb{E}} - \sum_{k=1}^{\mathcal{C}+1}y_{[k]}\log(h(g(x))_{[k]})
\end{equation}

$y \in \mathbb{R}^{\mathcal{C}+1}$ is the augmented label representation combining the closed and open classes simultaneously.

The cross-entropy loss $\mathcal{L}_{ce}$ can struggle with nuanced open and closed samples, resulting in less confidence for open sample predictions. To overcome this, we introduce an extra loss term targeting the entropy of predictions $h(g(x^{open}))$ for generated open samples $x^{open} \in \mathcal{X}^{open}$. This term aims to increase posterior probabilities for the synthesized open samples in the $\mathcal{C}+1^{th}$ class, while decreasing probabilities for closed class indices $1:\mathcal{C}$, ultimately enhancing confidence in open sample predictions. It's worth noting that $\mathcal{L}_{ce}$ already assigns these samples to the $\mathcal{C}+1^{th}$ index, and our approach further bolsters the confidence of this classification.
Furthermore, we introduce a margin loss for class-posterior probabilities in closed-set samples. This enhances prediction certainty and prevents finely distinguished closed samples from being misclassified as open-class data. Specifically, our aim is to widen the gap between the highest closed-set probability within indices $1: \mathcal{C}$ and the open-class probability (indexed as $\mathcal{C}+1$) for samples in $\mathcal{D}_s$.
These loss functions collectively constitute the discriminability objective ($\mathcal{L}_{disc}$), defined as follows, where $h(g(x))_{\mathcal{C}+1}$ and $h(g(x))_{top}$ represent posterior probabilities of the open class and the highest closed-set class, respectively, for a given input $x$.

\begin{equation}\label{Entropyeqn}
 \begin{split}
  \mathcal{L}_{disc}  = \underset{\mathcal{P}(\mathcal{X}^{open}, \{\mathcal{Y}^s \cup \mathcal{C}+1\})}{\mathbb{E}} - h(g(x)) \log(h(g(x))) & \\ - \underset{\mathcal{P}(\mathcal{X}^{s}, \mathcal{Y}^s)}{\mathbb{E}} |h(g(x^s))_{\mathcal{C}+1} - h(g(x^s))_{top }|_1^1 
\end{split}
\end{equation}

Finally, we introduce a novel margin loss ($\mathcal{L}_{sm}$) to ensure distinctiveness between the synthesized style properties $(\mu_{new}, \sigma_{new})$ and the input images' styles $(x_1^s, x_2^s)$. This objective encompasses separate losses for $\mu$ and $\sigma$, aiming to contain the generated $(\mu_{new}, \sigma_{new})$ within predefined bounds $(a,b)$ concerning the input styles. To achieve this, we consider two margins, striking a balance between dissimilarity from the inputs and preserving the image's semantic integrity.
In Eq. \ref{eq:Hingeeqn}, $||\mu_{new}-\mu_i||2$ is denoted as $d(\mu{new}, \mu_{i})$ for $i \in {1,2}$. This enforces the required distinction between the synthesized and input style properties.

\begin{equation}
\small
 \mathcal{L}_{sm}(\mu_{new},\mu_{i},a,b)  = \begin{cases}
 0 & \text{if $d(\mu_{new}, \mu_{i}) \in [a,b]$} \\
a - d(\mu_{new}, \mu_{i}) & \text{if $d(\mu_{new}, \mu_{i}) < a$} \\
d(\mu_{new}, \mu_{i}) - b & \text{if $d(\mu_{new}, \mu_{i}) > b$} \\
\end{cases}
\label{eq:Hingeeqn}
\end{equation}

% In order to generate distinctive novel styles, we incorporate a decorrelation criterion aimed at minimizing the pairwise cosine similarities of the generated styles. The updated $\mathcal{L}_{sm}$ can be represented as follows:

% \begin{equation}
%     \centering
%     \mathcal{L}_{sm} = \mathcal{L}_{sm} + \underset{m,n}{\sum} \left( \frac{\mu^m_{new} \cdot \mu^n_{new}}{|\mu^m_{new}||\mu^n_{new}|} + \frac{\sigma^m_{new} \cdot \sigma^n_{new}}{|\sigma^m_{new}||\sigma^n_{new}|} \right) 
%     \label{eq:sm1}
% \end{equation}

\noindent \textbf{Total loss and training:} To train the entire network in an end-to-end fashion, a weighted combination of losses is utilized, resulting in the total loss $\mathcal{L}_{T}$:

\begin{equation}\label{totalLosseqn}
 \mathcal{L}_{total}  = w_{ce}\mathcal{L}_{ce} + w_{disc}\mathcal{L}_{disc} + w_{sm}\mathcal{L}_{sm}   
\end{equation}

where $w_{ce}, w_{disc}, w_{sm}$ are the weights corresponding to the loss components and all are set to $1$ in our experiments.

\textbf{Testing:} At test time, the image is fed to the prediction model $g \circ h$ and the class label with the highest softmax probability score is predicted.

%\vspace*{-2mm}

\section{Experiments}\label{sec:exp}

\vspace*{-1mm}

\noindent \textbf{Datasets}:
Following the footsteps of \cite{12}, we conducted our experiments on four datasets: (1) \textbf{Office-31} \cite{13}, (2) \textbf{Digits} \cite{30, 31, 32, 33}, (3) \textbf{Office-Home} \cite{15}, and (4) \textbf{PACS} \cite{16}, respectively. The dataset details are mentioned in the \textsc{Supplementary}.

\noindent \textbf{Implementation details}:
While experimenting on \textit{Office31} and \textit{Digits} datasets, we use Amazon and MNIST as source domains while the rest are used as target domains, respectively. In the case of the rest of the two datasets, every domain serves as a source domain once, and the rest are treated as target domains. For all the datasets except Digits, we employ a \textsc{ResNet-18} pre-trained on ImageNet \cite{17} as the backbone network, whereas the \textsc{LeNet} \cite{lecun1998gradient} is considered for Digts. Intermediate feature maps for the domain synthesis task are extracted after the fifth convolutional block of the network. $\mathcal{F}_{ss}$ is realized through a fully connected neural network with an input layer of size 256. The input layer takes the concatenated values of $\mu_1, \sigma_1, \mu_2, \sigma_2$, with added noise from a standard normal distribution $\mathbb{N}(0, 1)$ to ensure style diversification. The input layer is followed by  a hidden layer with 192 nodes and an output layer of dimension 128. The first $64$ dimensions represent $\mu_{new}$ while the remaining half denotes $\sigma_{new}$. The ReLU activation function follows each of these layers. On the other hand, $\mathcal{F}_{fa}$ works as follows: First, it takes the input of concatenated features and projects it into a layer of dimension 512. This layer is followed by a ReLU activation and a Batch-Norm layer. The output of this layer is then passed to an output layer with a dimension of 512 and a Sigmoid activation.

The data is loaded in the form of triplets, where each consists of two images belonging to the same class and one randomly chosen image from a different class. The images belonging to the \textit{Digits} dataset are resized to the dimension of $28 \times 28$, while for all other datasets, the images are resized to $128 \times 128$. Normalization is performed using the ImageNet mean values of $[0.485, 0.456, 0.406]$ and standard deviation values of $[0.229, 0.224, 0.225]$ for the R-G-B channels, respectively. The data is loaded with a batch size of 160, and the SGD optimizer is used with a learning rate of 0.001 and a momentum of 0.9. During the training procedure, the triplets are shuffled every five epochs. The margin between the mean and standard deviation is kept within the ranges of $[1.5, 3.5]$ and $[0.1, 2]$, respectively. We provide the detailed architecture of our model, which had a total of $12\text{M}$ learnable parameters.%To implement the style diversification criterion of Eq. \ref{eq:sm1}, we maintain memory with the randomly selected set of generated styles and seek to update this memory periodically after every $10^{th}$ training iterations.

\noindent \textbf{Baselines and evaluation metrics}:
 In our study, we conducted a comparative analysis by benchmarking our results against various approaches. Firstly, we considered the Empirical Risk Minimization (ERM) method \cite{27} as a baseline without incorporating Single-DG. Additionally, we evaluated two state-of-the-art Single-DG methods: Adversarial Data Augmentation (ADA) \cite{volpi2018generalizing} and Maximum-Entropy Adversarial Data Augmentation (MEADA) \cite{26}. Furthermore, we assessed the performance of ERM, ADA, and MEADA after integrating with \textsc{CrossMatch} (represented as ``+CM") \cite{12}. To establish the baseline performance, we employed Open-Set Domain Adaptation by Back-propagation (OSDAP) \cite{28}, a prominent technique in open-set domain adaptation, as well as OpenMax \cite{29}, a method for open-set recognition.

During our experiments, we utilized several evaluation metrics, including overall accuracy ($acc$) and the h-score ($hs$) \cite{24}. The h-score represents the harmonic mean of the accuracy values for known and unknown classes, providing a comprehensive assessment by assigning a high score only when both accuracies are significantly high. We present the average performance, computed over three seeds, using the leave-one-domain-out approach. In this approach, we fix each domain as the source while evaluating the average performance over the remaining domains treated as targets.

\vspace*{-2mm}

\section{Results and Discussions}

\vspace*{-1mm}

\begin{table}[]\centering
\caption{Results (\% Accuracy) on \textbf{Office31} and \textbf{Digits} Dataset.}\label{tab: Office31-Digits}
\scriptsize
%\tiny
\vspace*{-3mm}
\begin{tabularx}{\linewidth}{l||YY|YY}\toprule
\multirow{2}{*}{Method} &\multicolumn{2}{c}{Office31} &\multicolumn{2}{c}{Digits} \\\cmidrule{2-5}
&$acc$ &$hs$ &$acc$ &$hs$ \\\midrule
OSDAP \cite{28} &76.51 &77.68 &41.42 &40.46 \\
OpenMax \cite{29} &18.19 &16.71 &42.38 &40.67 \\
ERM \cite{27} &79.82 &40.69 &49.17 &17.97 \\
ERM+CM \cite{12} &78.3 &51.14 &49.07 &40.15 \\
ADA \cite{volpi2018generalizing} &80.13 &38.65 &50.22 &20.14 \\
ADA+CM \cite{12} &78.61 &48.5 &49.71 &39.93 \\
MEADA \cite{26} &\textbf{80.26} &38.55 &52.98 &30.37 \\
MEADA+CM \cite{12} &78.98 &54.69 &51.27 &38.70 \\
\cmidrule{1-5}
\rowcolor[HTML]{d1ffff}\textsc{SODG-Net} &79.02&\textbf{78.62}&\textbf{55.66}  &\textbf{53.87} \\
\rowcolor[HTML]{d1ffff}\textsc{SODG-Net} $-\mathcal{L}_{disc}$ &78.34 &78.06  &53.64  &52.82 \\
\rowcolor[HTML]{d1ffff}\textsc{SODG-Net} $-\mathcal{L}_{disc} - \mathcal{L}_{sm}$ &75.88 &75.44  &53.26 &49.24 \\
\rowcolor[HTML]{d1ffff}\textsc{SODG-Net} $- \mathcal{L}_{sm}$ &68.66 &68.49  &47.80 &46.38 \\
\bottomrule
\end{tabularx}
\vspace*{-5mm}
\end{table}

%OfficeHome
\begin{table*}[htbp]\centering
\caption{Results (\% Accuracy) on \textbf{Office-Home} Dataset for different source domains.}\label{tab: Office-Home}
\scriptsize
%\tiny
\vspace*{-3mm}
\scalebox{0.9}{\begin{tabularx}{\textwidth}{l||YY|YY|YY|YY|YY}\toprule\multirow{2}{*}{Method} &\multicolumn{2}{c}{Art} &\multicolumn{2}{c}{Clipart} &\multicolumn{2}{c}{Product} &\multicolumn{2}{c}{Real-World} &\multicolumn{2}{c}{Average} \\
\cmidrule{2-11}
&$acc$ &$hs$ &$acc$ &$hs$ &$acc$ &$hs$ &$acc$ &$hs$ &$acc$ &$hs$ \\\midrule
OSDAP \cite{28} &45.61 &52.35 &52.78 &58.82 &41.45 &47.95 &53.51 &58.40 &48.34 &54.38 \\
OpenMax \cite{29} &22.42 &30.64 &22.67 &29.51 &15.10 &16.65 &25.54 &33.07 &21.43 &27.47 \\
ERM \cite{27} &65.00 &31.07 &64.12 &35.78 &60.53 &36.33 &66.59 &33.92 &64.06 &34.28 \\
ERM+CM \cite{12} &65.49 &52.85 &63.37 &50.51 &58.03 &47.25 &67.75 &52.60 &63.66 &50.80 \\
ADA \cite{volpi2018generalizing} &68.29 &32.94 &65.10 &42.09 &60.52 &34.72 &67.04 &34.86 &65.24 &36.15 \\
ADA+CM \cite{12} &66.30 &46.68 &62.64 &49.31 &58.72 &47.47 &66.82 &50.47 &63.62 &48.48 \\
MEADA \cite{26} &\textbf{68.31} &33.29 &65.25 &42.05 &60.43 &35.68 &67.04 &34.65 &65.01 &36.42 \\
MEADA+CM \cite{12} &65.85 &53.22 &62.90 &48.87 &58.36 &45.34 &67.10 &50.77 &63.55 &49.55 \\
\cmidrule{1-11}
\rowcolor[HTML]{d1ffff}\textsc{SODG-Net} &65.31 &\textbf{57.65} &\textbf{69.20} &\textbf{63.85} &\textbf{67.04} &\textbf{61.24} &\textbf{68.18} &\textbf{62.89} &\textbf{67.43} &\textbf{61.41} \\
\rowcolor[HTML]{d1ffff}\textsc{SODG-Net} $-\mathcal{L}_{disc}$  &65.03 &56.49 &64.81 &62.62 &65.99 &60.92 &67.82 &62.32 &65.91 &60.59 \\
\rowcolor[HTML]{d1ffff}\textsc{SODG-Net} $-\mathcal{L}_{disc} - \mathcal{L}_{sm}$ &58.67 &55.77 &60.66 &60.82 &64.08 &59.21 &62.65 &61.19 &61.51 &59.25 \\
\bottomrule
\end{tabularx}}
\vspace*{-3mm}
\end{table*}

%PACS
\begin{table*}[htbp]\centering
\caption{Results (\% Accuracy) on \textbf{PACS} Dataset for different source domains.}\label{tab: PACS}
\scriptsize
%\tiny
\vspace*{-3mm}
\scalebox{0.9}{\begin{tabularx}{\textwidth}{l||YY|YY|YY|YY|YY}\toprule

\multirow{2}{*}{Method} &\multicolumn{2}{c}{Art Painting} &\multicolumn{2}{c}{Cartoon} &\multicolumn{2}{c}{Sketch} &\multicolumn{2}{c}{Photo} &\multicolumn{2}{c}{Average} \\\cmidrule{2-11}
&$acc$ &$hs$ &$acc$ &$hs$ &$acc$ &$hs$ &$acc$ &$hs$ &$acc$ &$hs$ \\\midrule
OSDAP \cite{28} &53.30 &46.58 &43.73 &38.81 &42.05 &41.03 &30.81 &32.89 &42.47 &39.83 \\
OpenMax \cite{29} &52.59 &53.60 &31.71 &25.23 &29.85 &19.87 &27.60 &19.47 &35.44 &29.54 \\
ERM \cite{27} &62.24 &38.90 &55.34 &40.96 &39.19 &28.89 &38.32 &35.74 &48.77 &36.12 \\
ERM+CM \cite{12} &63.52 &44.90 &57.60 &48.31 &38.53 &30.43 &42.52 &41.60 &50.54 &41.31 \\
ADA \cite{volpi2018generalizing} &62.48 &39.02 &56.43 &41.55 &39.03 &26.93 &40.28 &38.13 &49.56 &36.41 \\
ADA+CM \cite{12} &\textbf{64.26} &42.40 &\textbf{60.41} &51.81 &42.48 &35.18 &43.97 &42.76 &52.78 &43.04 \\
MEADA \cite{26} &62.43 &38.85 &56.10 &41.34 &38.89 &26.43 &39.88 &38.24 &49.33 &36.22 \\
MEADA+CM \cite{12} &62.63 &41.88 &60.03 &51.36 &41.51 &35.76 &43.50 &41.60 &51.92 &42.65 \\
\cmidrule{1-11}
\rowcolor[HTML]{d1ffff}\textsc{SODG-Net} &57.02 &\textbf{57.44} &56.01 &\textbf{56.62} &\textbf{58.36} &\textbf{58.40} &\textbf{46.27} &\textbf{43.60} &\textbf{54.41} &\textbf{54.02} \\
\rowcolor[HTML]{d1ffff}\textsc{SODG-Net} $-\mathcal{L}_{disc}$ &56.41 &56.35 &55.41 &55.93 &58.12 &57.30 &43.07 &43.45 &53.25 &53.26 \\
\rowcolor[HTML]{d1ffff}\textsc{SODG-Net} $-\mathcal{L}_{disc} - \mathcal{L}_{sm}$ &55.33 &55.96 &50.66 &51.20 &57.56 &55.70 &40.52 &40.06 &51.02 &50.73 \\
\bottomrule

\end{tabularx}
}
\vspace*{-4mm}
\end{table*}

\begin{figure}
\begin{center}
\includegraphics[width = 0.75\linewidth]{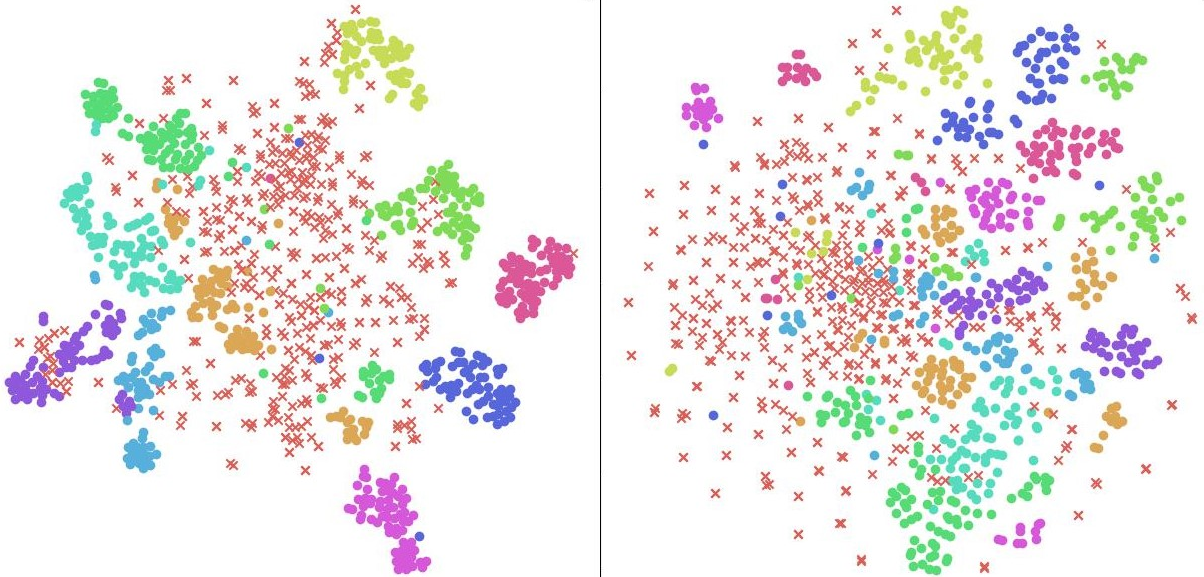}
\end{center}
\vspace*{-5mm}
   \caption{\textbf{t-SNEs on \textit{Office31} dataset}. Left t-SNE corresponds to the results obtained from \textsc{SODG-Net} and the right one to the network where the SSB is replaced with \textsc{MixStyle}. The red crosses `\textcolor{red}{$\times$}' in the central region represent the unknown class, while the clusters belong to the ten known classes from the source and the target domain, showing the overlap in the clusters.}
\label{fig:test_whole}
\vspace*{-5mm}
\end{figure}

\noindent\textbf{Office31} and \textbf{Digits}: Our experimental results on the \textit{Office31} and \textit{Digits} datasets, as presented in Table \ref{tab: Office31-Digits}, demonstrate the efficacy of our proposed architecture when compared to existing approaches in the field. In terms of the \textit{Office31} dataset, our method achieves performance on par with the highest accuracy ($acc$) reported by other methods. Notably, our approach exhibits an improvement of $0.94\%$ in the highest score ($hs$) over the OSDAP. Furthermore, our method surpasses MEADA by a considerable margin of $40.07\%$ in terms of $hs$. Regarding the \textit{Digits} dataset, the existing literature indicates that MEADA and OpenMax achieve the highest $acc$ and $hs$ scores, respectively. However, our \textsc{SODG-Net} outperforms these methods with a margin of $2.68\%$ in $acc$ and $13.2\%$ in the $hs$ score, respectively. In Fig. \ref{fig:test_whole}, it is evident that the source domain classes maintain a notable level of clustering even when faced with test data from previously unseen target domains. Additionally, the appearance of a distinct red cluster at the centre of the plot for the generated pseudo-classes confirms the effectiveness of our learning-driven open-sample synthesis scheme, which ensures that the generated open samples should lie outside the closed-space support. We can see that the network trained on closed samples generated using the \textsc{MixStyle} for SS-ODG underperforms 
 our \textsc{SODG-Net}. The cluseters in right t-SNE in Fig. \ref{fig:test_whole} are not as dense as the ones in the left t-SNE.

\noindent\textbf{Office-Home}: Table \ref{tab: Office-Home} showcases the results obtained on the challenging \textit{Office-Home} dataset. This dataset contains a larger number of classes compared to \textit{Office31} and \textit{Digits} and exhibits substantial distributional shifts between domains. Our proposed approach outperforms existing state-of-the-art methods in terms of $hs$, regardless of the chosen source domain. When considering the average performance across all domains, the ADA model achieves the highest $acc$, while OSDAP achieves the highest $hs$ in the literature. However, we achieve a remarkable improvement of $2.19\%$ over the best $acc$ achieved by previous methods and surpass the best $hs$ by an impressive margin of $7.03\%$.

\noindent\textbf{PACS}: The experimental results for the \textit{PACS} dataset are summarized in Table \ref{tab: PACS}. Similar to our performance on the \textit{Office-Home} dataset, our approach consistently outperforms the existing literature in terms of $hs$, regardless of the chosen source domain. When considering the average performances, our method achieves the highest accuracy ($acc$) and $hs$ scores, beating the next best technique combining ADA with \textsc{CrossMatch}  by $1.63\%$ and $10.98\%$, respectively, in the $acc$ and $hs$ metrics. The higher $hs$ is a clear indication of a better balance between the $acc_k$ and $acc_u$.

Furthermore, we show the sensitivity to the loss terms $\mathcal{L}_{sm}$ and $\mathcal{L}_{disc}$ for all the datasets, highlighting the importance of both the loss terms. We note that removing $\mathcal{L}_{sm}$ signifies an unbounded space for the generated styles. (see \textsc{supplementary} for $acc_k$ and $acc_u$)

\subsection{Ablation Study}
%\vspace*{-1mm}
\noindent \textbf{Diversity of the generated styles and open samples}: To quantitatively assess the diversity of the generated styles and open samples compared to the source domains, we compute the average cosine distance between the concatenated $(\mu, \sigma)$ of the source samples and the generated $(\mu_{new}, \sigma_{new})$. Similarly, we calculate this metric between the sets ${x^s}$ and ${x_{open}}$.
In the case of the \textit{Office31} dataset, we observe a mean cosine distance of $0.58$ and $0.72$ between the original and synthesized styles' mean and standard deviation, indicating significant separation in the embedding space. Likewise, the mean cosine distance between the closed and open features is $0.55$, further emphasizing their distinct placement.

\begin{figure}
\begin{center}
\includegraphics[width = 0.80\linewidth]{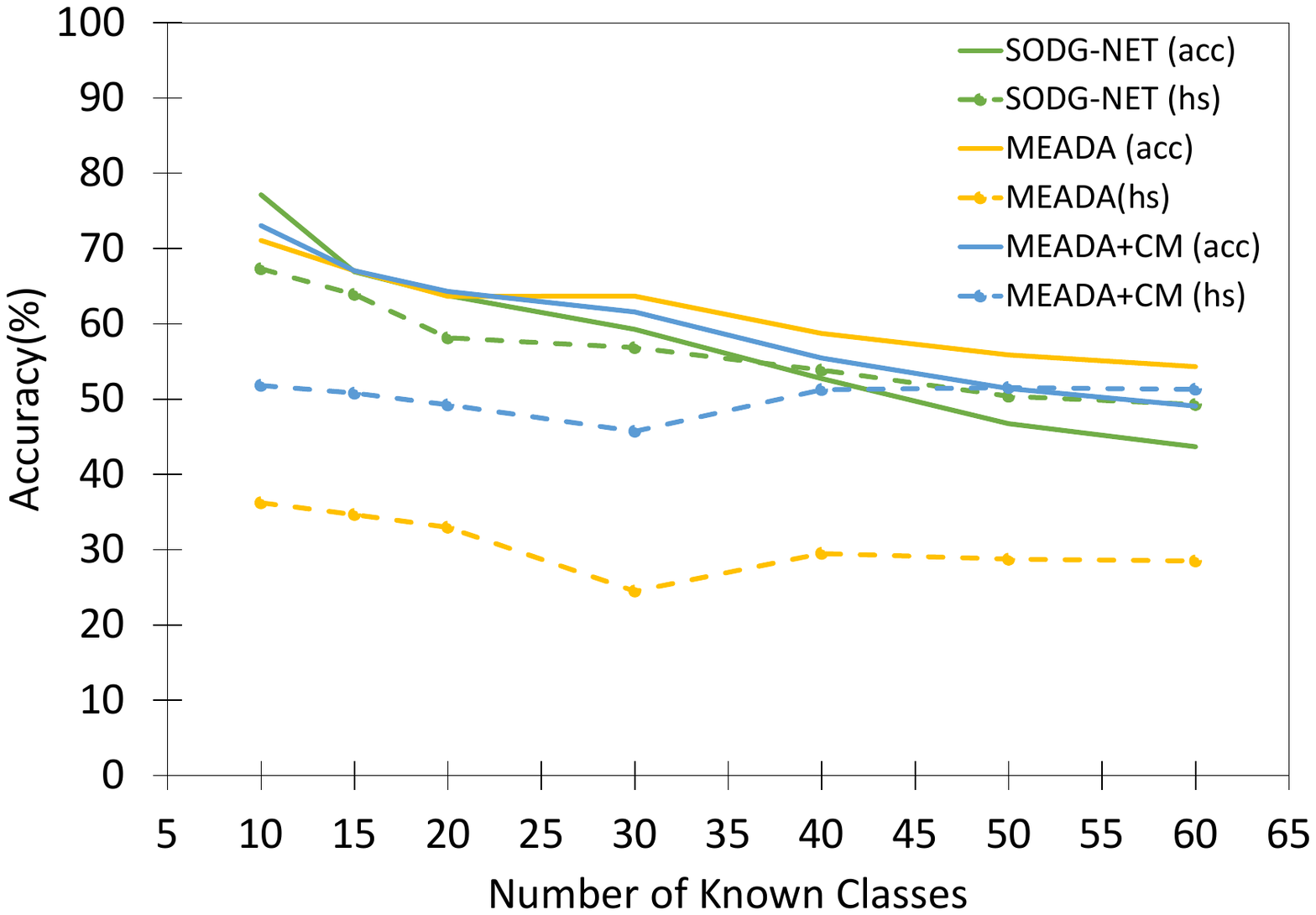}
\end{center}
\vspace*{-8mm}
   \caption{\textbf{Accuracy} vs \textbf{Number of Known Classes}. While performing similar to the MEADA and MEADA+CM in terms of $acc$, our \textsc{SODG-Net} outperforms the two methods for $hs$.}
\label{fig:known-class}
\vspace*{-2mm}
\end{figure}

\vspace{1.5mm}
\noindent\textbf{Varying the number of known classes}: We conducted experiments to evaluate the effectiveness of our method on the Real-World domain of the \textit{Office-Home} dataset, considering different $\mathcal{Y}^s$. The $|\mathcal{Y}^s|$ was varied from 10 to 60, incrementing in steps of $10$. Fig \ref{fig:known-class} illustrates the variation of $acc$ and $hs$ for the state-of-the-art MEADA method, the \textsc{CrossMatch} method applied over MEADA (MEADA+CM), and our proposed method. We observe that in our results, the $hs$ and $acc$ values are consistently close to each other, which again proves that our method has much more balance in $acc_k$ and $acc_u$ as compared to the other two. In particular, our $hs$ metric outperforms the other two methods when the number of known classes is smaller. As the number of known classes surpasses 50, our method yields comparable yet better results to MEADA+CM in terms of $hs$.

% \noindent\textbf{$\mathcal{F}_{ss}$ and MixStyle} \cite{zhou2021domain}: 
% The Mix-Style approach utilized a weight sampled from a Beta distribution to calculate $\mu_{new}$ and $\sigma_{new}$ values based on the mean and standard deviation of feature maps. In order to compare its performance with our learnable method, we conducted experiments by replacing $\mathcal{F}_{ss}$ with the Mix-Style approach. The results showed that the model using the Mix-Style approach under-performed compared to our learnable method, with a difference of $4.63\%$ and $4.84\%$ $acc$ and $hs$ respectively on \textit{Office31} dataset. See \textsc{Supplementary} for more details.

\begin{table}[]\centering
\caption{Ablation on different style augmentation techniques.}\label{tab: Office31-Digits-abl}
\scriptsize
%\tiny
\vspace*{-2mm}
\begin{tabularx}{\linewidth}{l||YY|YY}\toprule
\multirow{2}{*}{Method} &\multicolumn{2}{c}{Office31} &\multicolumn{2}{c}{Digits} \\ \cmidrule{2-5}
&$acc$ &$hs$ &$acc$ &$hs$ \\\midrule
Jin \etal \cite{jin2021style} &64.56 &58.43 &54.98 &40.33 \\
Kundu \etal \cite{kundu2021generalize} &58.98 &54.36 &51.27 &46.70 \\
Luo \etal \cite{luo2020adversarial} &64.58 &64.56 &54.05 &51.37\\
\cmidrule{1-5}
\rowcolor[HTML]{d1ffff}\textsc{SODG-Net} &\textbf{79.02} &\textbf{78.62} &\textbf{55.66} &\textbf{53.87} \\
% \rowcolor[HTML]{d1ffff}\textsc{SODG-Net} $- \mathcal{L}_{sm}$ &68.66 &68.49  &47.80 &46.38 \\
\bottomrule
\end{tabularx}
\vspace*{-3mm}
\end{table}

\vspace{1.5mm}
\noindent \textbf{Comparison of the style synthesis approaches:} In Table \ref{tab: Office31-Digits-abl}, we conduct a comparative analysis of the style synthesis block within \textsc{SODG-Net} against three techniques from the DG literature \cite{jin2021style,kundu2021generalize,luo2020adversarial}. These methods rely on feature statistics interpolation to create new domains. The results unequivocally demonstrate the superior performance of \textsc{SODG-Net} across both closed and open-set scenarios. This substantial advantage can be attributed to our metric-driven approach that fosters diversified style generation, profoundly enhancing the model's overall generalizability.

\vspace{1.5mm}
\noindent\textbf{$\mathcal{F}_{fa}$ and mixup based open sample generation methods}: 
In order to generate pseudo-samples for open-set recognition, we employed $\mathcal{F}_{fa}$ that learns a weight to linearly combine the content features obtained from the backbone network. As an alternative approach, we conducted experiments on \textit{Office31} dataset using the following samples as unknown classes: (1) Taking two images from different classes, cropping them in half, and joining them together \cite{mancini2020towards}, (2) Calculating the mean of corresponding pixel values in images from two different classes, and (3) Randomly replacing a $30\times30$ patch from one image with a patch of the same size from an image of another class.
When comparing these alternative methods with our proposed approach of generating open-set representations, we observed that the first method resulted in a performance lag of $4.22\%$ and $6.04\%$ for accuracy and harmonic mean score, respectively. The other two methods were unable to generalize effectively on known classes, consequently failing to achieve consistent training accuracy.

\vspace*{-1mm}

\begin{table}[!htp]\centering
\caption{Effects of variations in the interval $[a, b]$ on $acc$. }\label{tab: acc}
\vspace*{-3mm}
\scriptsize
%\tiny
\begin{tabularx}{\linewidth}{YY|YYYYYYY}\toprule
\multicolumn{2}{c}{\multirow{2}{*}{}} &\multicolumn{5}{c}{Standard Deviation} \\\cmidrule{3-7}
& &[0.1,1] &[1,2] &[2,3] &[3,4] &[4,5] \\\midrule
\multirow{5}{*}{Mean} &[0.1,1] &77.52 &75.89 &76.7 &78.61 &76.98 \\
&[1,2] &76.7 &76.7 &76.16 &76.84 &77.38 \\
&[2,3] &75.34 &76.98 &77.38 &77.66 &75.07 \\
&[3,4] &75.48 &77.25 &76.84 &74.48 &75.07 \\
&[4,5] &76.84 &77.25 &76.43 &75.07 &75.75 \\
\bottomrule
\end{tabularx}
\vspace*{-5mm}
\end{table}

\begin{table}[!htp]\centering
\caption{Effects of variations in the interval $[a, b]$ on $hs$. }\label{tab: hs}
\vspace*{-3mm}
\scriptsize
%\tiny
\begin{tabularx}{\linewidth}{YY|YYYYYYY}\toprule
\multicolumn{2}{c}{\multirow{2}{*}{}} &\multicolumn{5}{c}{Standard Deviation} \\\cmidrule{3-7}
& &[0.1,1] &[1,2] &[2,3] &[3,4] &[4,5] \\\midrule
\multirow{5}{*}{Mean} &[0.1,1] &76.71 &75.45 &75.99 &77.17 &76.9 \\
&[1,2] &75.94 &76.23 &75.87 &75.75 &76.98 \\
&[2,3] &75.24 &76.62 &76.6 &77.46 &74.78 \\
&[3,4] &75.1 &76.06 &73.94 &73.83 &74.97 \\
&[4,5] &76.47 &75.25 &76.37 &74.49 &74.5 \\
\bottomrule
\end{tabularx}
\vspace*{-2mm}
\end{table}

\noindent\textbf{Effects of $a$ and $b$ in $\mathcal{L}_{sm}$}: In order to produce different styles, we induce a lower bound denoted as $a$ to ensure a minimum distance between the predicted statistical features and the original style primitives. Simultaneously, an upper bound denoted as $b$ is applied to prevent significant alterations in the distribution that could result in changes to the semantic information of the feature map. The experimentation involved evaluating five different ranges of $[a, b]$ for both the mean and standard deviation, resulting in a total of 25 combinations. The experimental results, presented in tables \ref{tab: acc} and \ref{tab: hs}, provide insights into the $acc$ and $hs$ achieved on the \textit{Office31} dataset.

\vspace*{-2mm}

\section{Takeaways}

\vspace*{-1mm}

This paper addresses the challenge of single-source ODG by proposing an architecture called \textsc{SODG-Net}. Our goal is to develop a model that can effectively generalize to diverse target domains using data from a single source domain, while also accommodating unknown classes in the target domain. To achieve this, we propose novel losses that enable learning to synthesize statistical style features different from the source domain, facilitating effective generalization for known classes. Additionally, \textsc{SODG-Net} generates representations for unknown classes by combining content features from images of two distinct classes using a learnable feature weighting and a classification constraint. Extensive experimentation on various datasets validates that our \textsc{SODG-Net} significantly improves generalization to different target domains using only a single source domain, while effectively detecting unknown classes. Future directions include applying this model to other application domains such as person re-identification.

{\small
\bibliographystyle{ieee_fullname}
\bibliography{egbib}
}

\end{document}

%%%%%%%%% TITLE
\title{SUPPLEMENTARY MATERIAL \\ Learning Class and Domain Augmentations for Single-Source Open-Domain Generalization}

% \author{Prathmesh Bele\\
% Indian Institute of Technology Bombay\\
% Mumbai, India - 400076\\
% {\tt\small prathmeshbele.iitb@gmail.com}
% \and
% Avigyan Bhattacharya\\
% Jadavpur University\\
% Kolkata, India - 700032\\
% {\tt\small avigyanbhattacharya123@gmail.com}
% }

\author{Prathmesh Bele$^{1}$ \quad \hspace{-0.45cm} Valay Bundele$^{1}\thanks{equal contribution}$ \quad \hspace{-0.4cm} Avigyan Bhattacharya$^{1*}$ \quad \hspace{-0.45cm} Ankit Jha$^{1}$ \quad \hspace{-0.45cm} Gemma Roig$^{2}$ \quad \hspace{-0.45cm} Biplab Banerjee$^{1}$ \vspace{0cm} \\
{\normalsize $^1$Indian Institute of Technology Bombay, India} \quad
{\normalsize $^2$Goethe University Frankfurt, Germany} \quad \\
{\tt \small {prathmeshbele.iitb@gmail.com, valaybundele@gmail.com, avigyanbhattacharya123@gmail.com}}\\
{\tt \small{ankitjha16@gmail.com, roig@cs.uni-frankfurt.de, getbiplab@gmail.com}}
}

\maketitle
\thispagestyle{empty}

In this supplementary material, we present a comprehensive dataset description, elaborate model architecture, algorithm details, and tables showcasing the accuracies of both known and unknown classes obtained from our experiments.

% \begin{figure*}
% \begin{center}
% \includegraphics[width = 0.7\linewidth]{images/supp_mix.jpg}
% \end{center}
% \vspace*{-5.5mm}
%    \caption{Samples of images which were cropped in half and concatenated with an image from another class to act as representatives of unknown class samples}
% \label{fig:test_whole}

% \end{figure*}

\section{Dataset Description}
(1) \textbf{Office-31} \cite{13}: This dataset comprises 31 classes obtained from three distinct domains: Amazon, DSLR, and Webcam, totaling 4652 images. In our experiments, we consider the 10 classes shared by Office-31 and Caltech-256 \cite{14} (backpack, bike, calculator, headphones, keyboard, laptop, monitor, mouse, mug, and projector) as the source domain label space, following the suggestion of \cite{Busto_2017_ICCV}. The remaining eleven classes in alphabetical order (ruler, punchers, stapler, scissors, trash can, tape dispenser, pen, phone, printer, ring binder, and speaker) constitute the target unknown class space. Due to the relatively fewer number of samples in the DSLR and Webcam domains, we conduct experiments solely on the Amazon domain as the source domain.

(2) \textbf{Digits}: This dataset consists of five digit datasets: \textbf{MNIST} \cite{30}, \textbf{SVHN} \cite{31}, \textbf{USPS} \cite{32}, \textbf{MNIST-M}, and \textbf{SYN} \cite{33}. In our setup, \textbf{MNIST} serves as the source domain with known classes representing numbers from 0 to 4, while the other datasets are considered as target domains, representing unknown classes for numbers 5 to 9. We select 10,000 images from the MNIST dataset, following the approach of \cite{volpi2018generalizing} and \cite{26}, for the source domain.

(3) \textbf{Office-Home} \cite{15}: This dataset comprises data from four different domains: Art, Clipart, Product, and Real-World, totaling 15,500 images. Each domain consists of 65 classes, with the first 15 classes (alarm clock, backpack, battery, bed, bike, bottle, bucket, calculator, calendar, candles, chair, clipboards, computer, couch, and curtains) used as the source label space, while the remaining 50 classes are considered as unknown target classes.

(4) \textbf{PACS} \cite{16}: This dataset contains 9,991 images from four domains: Art Painting, Cartoon, Photo, and Sketch. Each domain includes images from seven different classes. In our setup, we utilize four classes (dog, elephant, giraffe, and guitar) as the label space in the source domain, while the remaining three classes (horse, house, and person) are treated as unknown classes in the target domains.

\section{Model Architecture}
The detailed architectures of each of the networks the style synthesis block $\mathcal{F}_{ss}$, the feature aggregation block $\mathcal{F}_{fa}$, $g^l$, $g^{l+1:L}$ are given in the tables  \ref{tab:Fss-arc}, \ref{tab:Ffa-arc}, \ref{tab:gl-arc}, \ref{tab:gL-arc},  respectively. In our experiments, we have taken the encoder network to be \textsc{ResNet18} and $g^l$ consists of first five convolutional blocks while $g^{l+1:L}$ is rest of the network i.e., \textsc{ResNet18} $ = g^{l+1:L} \circ g^{l}$. We also conduct experiments by changing the point of extraction of feature map to apply the style synthesis block. Table \ref{tab:gl-arc-small} represents the architecture of $g^l$ which was shallower as compared to the one in \ref{tab:gl-arc} while table \ref{tab:gl-arc-large} shows the deeper version. In both the cases, rest of the \textsc{ResNet18} was used as $g^{l+1:L}$. Results of these experiments are given in table \ref{tab: diff-depth}. The `$-1$' in output shapes is a placeholder for the batch-size.

\begin{table}[h]
\centering
\scriptsize
\caption{Architecture Summary of $\mathcal{F}_{ss}$}
\begin{tabularx}{\linewidth}{Y|Y|Y}
\hline
\textbf{Layer (type)} & \textbf{Output Shape} & \textbf{Param \#} \\
\hline
    \hline
    Input Shape & $[-1, 256]$ & 0 \\ 
Linear-1 & $[-1, 192]$ & 49,344 \\
ReLU-2 & $[-1, 192]$ & 0 \\
Linear-3 &$ [-1, 128]$ & 24,704 \\
ReLU-4 & $[-1, 128]$ & 0 \\
\hline
\multicolumn{3}{|c|}{Total parameters: 74,048} \\
\multicolumn{3}{|c|}{Trainable parameters: 74,048} \\
\multicolumn{3}{|c|}{Non-trainable parameters: 0} \\
\hline
\end{tabularx}
\label{tab:Fss-arc}
\end{table}
%%%%%%%%%%%%%%%%%%%%%%%%%%%%%%%%%%%%%%%%%%%%%%%%%%%%%%%%%%%%%

\begin{table}[htbp]
  \centering
  \scriptsize
  \caption{Architecture Summary of $\mathcal{F}_{fa}$}
  \label{tab:Ffa-arc}
\begin{tabularx}{\linewidth}{Y|Y|Y}
    \hline
    \textbf{Layer (type)} & \textbf{Output Shape} & \textbf{Param \#} \\
    \hline
    \hline
    Input Shape & $[-1, 1024]$ & 0 \\
    Linear-1 & $[-1, 512]$ & 524,800 \\
    ReLU-2 & $[-1, 512]$ & 0 \\
    BatchNorm1d-3 & $[-1, 512]$ & 1,024 \\
    Linear-4 & $[-1, 512]$ & 262,656 \\
    Sigmoid-5 & $[-1, 512]$ & 0 \\
    \hline
\multicolumn{3}{|c|}{Total parameters: 788,480} \\
\multicolumn{3}{|c|}{Trainable parameters: 788,480} \\
\multicolumn{3}{|c|}{Non-trainable parameters: 0} \\
    \hline
  \end{tabularx}
\end{table}
%%%%%%%%%%%%%%%%%%%%%%%%%%%%%%%%%%%%%%%%%%%%%%%%%%%%%%%%%%%%%

\begin{table}[h]
\centering
\scriptsize
\caption{Architecture Summary of $g^l$}
\label{tab:gl-arc}
\begin{tabularx}{\linewidth}{Y|Y|Y}
\hline
\textbf{Layer (type)} & \textbf{Output Shape} & \textbf{Param \#} \\
\hline
\hline
Input Shape & $[-, 3, 128, 128]$ & 0 \\
Conv2d-1 & $[-1, 64, 64, 64]$ & 9,408 \\
BatchNorm2d-2 & $[-1, 64, 64, 64]$ & 128 \\
ReLU-3 & $[-1, 64, 64, 64]$ & 0 \\
MaxPool2d-4 & $[-1, 64, 32, 32]$ & 0 \\
Conv2d-5 & $[-1, 64, 32, 32]$ & 36,864 \\
BatchNorm2d-6 & $[-1, 64, 32, 32]$ & 128 \\
ReLU-7 & $[-1, 64, 32, 32]$ & 0 \\
Conv2d-8 & $[-1, 64, 32, 32]$ & 36,864 \\
BatchNorm2d-9 & $[-1, 64, 32, 32]$ & 128 \\
ReLU-10 & $[-1, 64, 32, 32]$ & 0 \\
BasicBlock-11 & $[-1, 64, 32, 32]$ & 0 \\
Conv2d-12 & $[-1, 64, 32, 32]$ & 36,864 \\
BatchNorm2d-13 & $[-1, 64, 32, 32]$ & 128 \\
ReLU-14 & $[-1, 64, 32, 32]$ & 0 \\
Conv2d-15 & $[-1, 64, 32, 32]$ & 36,864 \\
BatchNorm2d-16 & $[-1, 64, 32, 32]$ & 128 \\
ReLU-17 & $[-1, 64, 32, 32]$ & 0 \\
BasicBlock-18 & $[-1, 64, 32, 32]$ & 0 \\
\hline
\multicolumn{3}{|c|}{Total parameters: 157,504} \\
\multicolumn{3}{|c|}{Trainable parameters: 157,504} \\
\multicolumn{3}{|c|}{Non-trainable parameters: 0} \\
\hline
\end{tabularx}
\end{table}
%%%%%%%%%%%%%%%%%%%%%%%%%%%%%%%%%%%%%%%%%%%%%%%%%%%%%%%%%%%%%%
\begin{table}[htbp]
  \centering
  \scriptsize
  \caption{Architecture Summary of $g^{l+1:L}$}
  \label{tab:gL-arc}
\begin{tabularx}{\linewidth}{Y|Y|Y}
    \hline
    \textbf{Layer (type)} & \textbf{Output Shape} & \textbf{Param \#} \\
    \hline
    \hline
    Input Shape & $[-1, 64, 32, 32]$ & 0 \\
    Conv2d-1 & $[-1, 128, 16, 16]$ & 73,728 \\
BatchNorm2d-2 &$ [-1, 128, 16, 16]$ & 256 \\
ReLU-3 & $[-1, 128, 16, 16]$ & 0 \\
Conv2d-4 & $[-1, 128, 16, 16]$ & 147,456 \\
BatchNorm2d-5 & $[-1, 128, 16, 16]$ & 256 \\
Conv2d-6 & $[-1, 128, 16, 16]$ & 8,192 \\
BatchNorm2d-7 & $[-1, 128, 16, 16]$ & 256 \\
ReLU-8 & $[-1, 128, 16, 16]$ & 0 \\
BasicBlock-9 &$ [-1, 128, 16, 16]$ & 0 \\
Conv2d-10 & $[-1, 128, 16, 16]$ & 147,456 \\
BatchNorm2d-11 & $[-1, 128, 16, 16]$ & 256 \\
ReLU-12 & $[-1, 128, 16, 16]$ & 0 \\
Conv2d-13 & $[-1, 128, 16, 16]$ & 147,456 \\
BatchNorm2d-14 & $[-1, 128, 16, 16]$ & 256 \\
ReLU-15 & $[-1, 128, 16, 16]$ & 0 \\
BasicBlock-16 & $[-1, 128, 16, 16]$ & 0 \\
Conv2d-17 & $[-1, 256, 8, 8]$ & 294,912 \\
BatchNorm2d-18 & $[-1, 256, 8, 8]$ & 512 \\
ReLU-19 & $[-1, 256, 8, 8]$ & 0 \\
Conv2d-20 & $[-1, 256, 8, 8]$ & 589,824 \\
BatchNorm2d-21 & $[-1, 256, 8, 8]$ & 512 \\
Conv2d-22 & $[-1, 256, 8, 8]$ & 32,768 \\
BatchNorm2d-23 & $[-1, 256, 8, 8]$ & 512 \\
ReLU-24 & $[-1, 256, 8, 8]$ & 0 \\
BasicBlock-25 & $[-1, 256, 8, 8]$ & 0 \\
Conv2d-26 & $[-1, 256, 8, 8]$ & 589,824 \\
BatchNorm2d-27 & $[-1, 256, 8, 8]$ & 512 \\
ReLU-28 & $[-1, 256, 8, 8]$ & 0 \\
Conv2d-29 & $[-1, 256, 8, 8]$ & 589,824 \\
BatchNorm2d-30 & $[-1, 256, 8, 8]$ & 512 \\
ReLU-31 & $[-1, 256, 8, 8]$ & 0 \\
BasicBlock-32 & $[-1, 256, 8, 8]$ & 0 \\
Conv2d-33 & $[-1, 512, 4, 4]$ & 1,179,648 \\
BatchNorm2d-34 & $[-1, 512, 4, 4]$ & 1,024 \\
ReLU-35 & $[-1, 512, 4, 4]$ & 0 \\
Conv2d-36 & $[-1, 512, 4, 4]$ & 2,359,296 \\
BatchNorm2d-37 & $[-1, 512, 4, 4]$ & 1,024 \\
Conv2d-38 & $[-1, 512, 4, 4]$ & 131,072 \\
BatchNorm2d-39 & $[-1, 512, 4, 4]$ & 1,024 \\
ReLU-40 & $[-1, 512, 4, 4]$ & 0 \\
BasicBlock-41 & $[-1, 512, 4, 4]$ & 0 \\
Conv2d-42 & $[-1, 512, 4, 4]$ & 2,359,296 \\
BatchNorm2d-43 & $[-1, 512, 4, 4]$ & 1,024 \\
ReLU-44 & $[-1, 512, 4, 4]$ & 0 \\
Conv2d-45 & $[-1, 512, 4, 4]$ & 2,359,296 \\
BatchNorm2d-46 & $[-1, 512, 4, 4]$ & 1,024 \\
ReLU-47 & $[-1, 512, 4, 4]$ & 0 \\
BasicBlock-48 & $[-1, 512, 4, 4]$ & 0 \\
AdaptiveAvgPool2d-49 & $[-1, 512, 1, 1$] & 0 \\
Identity-50 & $[-1, 512, 1, 1]$ & 0 \\
    \hline
\multicolumn{3}{|c|}{Total parameters: 11,019,008} \\
\multicolumn{3}{|c|}{Trainable parameters: 11,019,0080} \\
\multicolumn{3}{|c|}{Non-trainable parameters: 0} \\
    \hline
  \end{tabularx}
\end{table}

%%%%%%%%%%%%%%%%%%%%%%%%%%%%%%
\begin{table}[h]
\centering
\scriptsize
\caption{Architecture Summary of shallow $g^l$}
\label{tab:gl-arc-small}
\begin{tabularx}{\linewidth}{Y|Y|Y}
\hline
\textbf{Layer (type)} & \textbf{Output Shape} & \textbf{Param \#} \\
\hline
\hline
Input Shape & $[-, 3, 128, 128]$ & 0 \\
Conv2d-1 & $[-1, 64, 64, 64]$ & 9,408 \\
BatchNorm2d-2 & $[-1, 64, 64, 64]$ & 128 \\
ReLU-3 & $[-1, 64, 64, 64]$ & 0 \\
MaxPool2d-4 & $[-1, 64, 32, 32]$ & 0 \\
\hline
\multicolumn{3}{|c|}{Total parameters: 9,536} \\
\multicolumn{3}{|c|}{Trainable parameters: 9,536} \\
\multicolumn{3}{|c|}{Non-trainable parameters: 0} \\
\hline
\end{tabularx}
\end{table}

%%%%%%%%%%%%%%%%%%%%%%%%%%%%%
\begin{table}[h]
\centering
\scriptsize
\caption{Architecture Summary of deep $g^l$}
\label{tab:gl-arc-large}
\begin{tabularx}{\linewidth}{Y|Y|Y}
\hline
\textbf{Layer (type)} & \textbf{Output Shape} & \textbf{Param \#} \\
\hline
\hline
Input Shape & $[-, 3, 128, 128]$ & 0 \\
Conv2d-1 & $[-1, 64, 64, 64]$ & 9,408 \\
BatchNorm2d-2 & $[-1, 64, 64, 64]$ & 128 \\
ReLU-3 & $[-1, 64, 64, 64]$ & 0 \\
MaxPool2d-4 & $[-1, 64, 32, 32]$ & 0 \\
Conv2d-5 & $[-1, 64, 32, 32]$ & 36,864 \\
BatchNorm2d-6 & $[-1, 64, 32, 32]$ & 128 \\
ReLU-7 & $[-1, 64, 32, 32]$ & 0 \\
Conv2d-8 & $[-1, 64, 32, 32]$ & 36,864 \\
BatchNorm2d-9 & $[-1, 64, 32, 32]$ & 128 \\
ReLU-10 & $[-1, 64, 32, 32]$ & 0 \\
BasicBlock-11 & $[-1, 64, 32, 32]$ & 0 \\
Conv2d-12 & $[-1, 64, 32, 32]$ & 36,864 \\
BatchNorm2d-13 & $[-1, 64, 32, 32]$ & 128 \\
ReLU-14 & $[-1, 64, 32, 32]$ & 0 \\
Conv2d-15 & $[-1, 64, 32, 32]$ & 36,864 \\
BatchNorm2d-16 & $[-1, 64, 32, 32]$ & 128 \\
ReLU-17 & $[-1, 64, 32, 32]$ & 0 \\
BasicBlock-18 & $[-1, 64, 32, 32]$ & 0 \\
Conv2d-19 & $[-1, 128, 16, 16]$ & 73,728 \\
BatchNorm2d-20 & $[-1, 128, 16, 16]$ & 256 \\
ReLU-21 & $[-1, 128, 16, 16]$ & 0 \\
Conv2d-22 & $[-1, 128, 16, 16]$ & 147,456 \\
BatchNorm2d-23 & $[-1, 128, 16, 16]$ & 256 \\
Conv2d-24 & $[-1, 128, 16, 16]$ & 8,192 \\
BatchNorm2d-25 & $[-1, 128, 16, 16]$ & 256 \\
ReLU-26 & $[-1, 128, 16, 16]$ & 0 \\
BasicBlock-27 & $[-1, 128, 16, 16]$ & 0 \\
Conv2d-28 & $[-1, 128, 16, 16]$ & 147,456 \\
BatchNorm2d-29 & $[-1, 128, 16, 16]$ & 256 \\
ReLU-30 & $[-1, 128, 16, 16]$ & 0 \\
Conv2d-31 & $[-1, 128, 16, 16]$ & 147,456 \\
BatchNorm2d-32 & $[-1, 128, 16, 16]$ & 256 \\
ReLU-33 & $[-1, 128, 16, 16]$ & 0 \\
BasicBlock-34 & $[-1, 128, 16, 16]$ & 0 \\
\hline
\multicolumn{3}{|c|}{Total parameters: 683,072} \\
\multicolumn{3}{|c|}{Trainable parameters: 683,072} \\
\multicolumn{3}{|c|}{Non-trainable parameters: 0} \\
\hline
\end{tabularx}
\end{table}

%%%%%%%%%%%%%%%%%%%%%%%%%
\begin{table}[!htp]\centering
\caption{Results on different depths of $g^l$ on Office31 dataset}\label{tab: diff-depth}
\scriptsize
\begin{tabularx}{\linewidth}{Y||Y|Y|Y}\toprule
Metric &Shallow $g^l$ Table \ref{tab:gl-arc-small} &Our $g^l$ Table \ref{tab:gl-arc} &Deeper $g^l$ Table \ref{tab:gl-arc-large}\\\midrule

$acc_k$ &70.64 &\textbf{73.96} &75.35 \\
$acc_u$ &60.59 &\textbf{83.91} &66.76 \\
$acc$ &65.53 &\textbf{79.02} &70.98 \\
$hs$ &65.23 &\textbf{78.62} &70.79 \\
\bottomrule
\end{tabularx}
\end{table}
 
%%%%%%%%%%%%%%%%%%%%%%%%%%%%%%%%%%%%%%%%%%%%%%%%%%%%%%%%%%%%%
\section{Experiments on the large scale dataset DomainNet}
The table \ref{tab: DomainNet} contains the results on \textit{DomainNet} dataset. We had chosen four domains out of six from the dataset (Clipart, Painting, Sketch and Real). For our experiments, we selected alphabetically first 150 classes from each of the four domains and remaining 195 classes were treated as unknown target class. The total number of samples corresponding to these four domain is 362470. We compare our results against two bechmark methods, ERM \cite{27} and ADA \cite{volpi2018generalizing}. Here, we outperform the ADA by $11.17\%$ for average accuracy ($acc$) and by $3.15\%$ while compared to the h-score ($hs$).

\section{Experimental Results with Known and Unknown Class Accuracies}

In this section we report the known and unknown class accuracies ($acc_k$ and $acc_u$) for the experiments conducted. The table \ref{tab: Office31-Digits-uk} has the results for \textit{Office31} and \textit{Digits} datasets while table \ref{tab: Office-Home-uk} and \ref{tab: PACS-uk} have the results for \textit{Office-Home} and \textit{PACS} datasets, respectively.

\begin{table}[htbp]\centering
\caption{$acc_k$ \& $acc_u$ (\% Accuracy) on \textbf{Office31} and \textbf{Digits} Dataset.}\label{tab: Office31-Digits-uk}
\scriptsize
\begin{tabularx}{\linewidth}{l||YY|YY}\toprule
\multirow{2}{*}{Method} &\multicolumn{2}{c}{Office31} &\multicolumn{2}{c}{Digits} \\\cmidrule{2-5}
&$acc_k$ &$acc_u$ &$acc_k$ &$acc_u$ \\\midrule
OSDAP \cite{28}  &75.77 &84.28 &35.59 &70.60 \\
OpenMax \cite{29} &10.01 &100 &34.40 &83.81  \\
ERM \cite{27} &85.1 &27.04 &56.40 &13.04 \\
ERM+CM \cite{12} &82.37 &37.6 &48.67 &53.52 \\
ADA \cite{volpi2018generalizing} &85.62 &25.24 &57.24 &15.11 \\
ADA+CM \cite{12} &53.02 &34.51 &49.24 &52.07 \\
MEADA \cite{26} &85.78 &25.09 &57.61 &29.83 \\
MEADA+CM \cite{12} &82.77 &41.08 &52.30 &46.11 \\
\cmidrule{1-5}
\rowcolor[HTML]{d1ffff}\textsc{SODG-NET} &73.96 &83.91 &43.60 &70.45 \\
\rowcolor[HTML]{d1ffff}\textsc{SODG-NET} $-\mathcal{L}_{disc}$ &74.24 &82.31 &44.63 &64.71 \\
\rowcolor[HTML]{d1ffff}\textsc{SODG-NET} $-\mathcal{L}_{disc} - \mathcal{L}_{sm}$ &73.13 &78.55 &41.97 &67.11 \\
\rowcolor[HTML]{d1ffff}\textsc{SODG-Net $- \mathcal{L}_{sm}$} &{65.65} &{71.58}  &{37.76} &{60.12} \\
\bottomrule
\end{tabularx}
\end{table}

%OfficeHome
\begin{table*}[htbp]\centering
\caption{$acc_k$ \& $acc_u$ (\% Accuracy) on \textbf{Office-Home} Dataset.}\label{tab: Office-Home-uk}
\scriptsize
\begin{tabularx}{\textwidth}{l||YY|YY|YY|YY|YY}\toprule\multirow{2}{*}{Method} &\multicolumn{2}{c}{Art} &\multicolumn{2}{c}{Clipart} &\multicolumn{2}{c}{Product} &\multicolumn{2}{c}{Real-World} &\multicolumn{2}{c}{Average} \\
\cmidrule{2-11}
&$acc_k$ &$acc_u$ &$acc_k$ &$acc_u$ &$acc_k$ &$acc_u$  &$acc_k$ &$acc_u$  &$acc_k$ &$acc_u$  \\\midrule
OSDAP \cite{28} &44.13 &67.84 &51.69 &69.26 &40.00 &63.47 &52.48 &66.92 &47.07 &66.87 \\
OpenMax \cite{29} &17.38 &98.08 &17.72 &97.04 &9.53 &98.59 &20.78 &97.01 &16.35 &97.68 \\
ERM \cite{27} &68.54 &20.53 &66.75 &24.65 &62.81 &26.26 &69.48 &23.18 &66.90 &23.66 \\
ERM+CM \cite{12} &66.48 &48.57 &64.80 &41.95 &59.17 &40.94 &69.36 &43.69 &64.95 &43.79 \\
ADA \cite{volpi2018generalizing} &71.36 &22.05 &67.37 &31.19 &62.91 &24.55 &69.92 &23.88 &67.89 &25.42 \\
ADA+CM \cite{12} &67.53 &39.59 &64.10 &40.67 &59.92 &40.72 &68.53 &40.79 &65.02 &40.44 \\
MEADA \cite{26} &71.37 &22.36 &66.45 &31.27 &62.75 &25.60 &69.92 &23.71 &67.62 &25.74 \\
MEADA+CM \cite{12} &66.63 &45.28 &64.43 &37.84 &59.74 &37.71 &68.82 &41.28 &64.90 &40.53 \\
\cmidrule{1-11}
\rowcolor[HTML]{d1ffff}\textsc{SODG-NET} &48.33 &71.41 &56.23 &73.86 &53.28 &71.98 &55.38 &72.77 &53.31 &72.50 \\
\rowcolor[HTML]{d1ffff}\textsc{SODG-NET} $-\mathcal{L}_{disc}$  &46.62 &71.64 &58.81 &66.96 &53.47 &70.49 &54.61 &72.57 &53.38 &70.41 \\
\rowcolor[HTML]{d1ffff}\textsc{SODG-NET} $-\mathcal{L}_{disc} - \mathcal{L}_{sm}$ &51.09 &61.39 &61.18 &60.47 &52.24 &68.33 &58.52 &64.13 &55.76 &63.58 \\
\bottomrule
\end{tabularx}
\end{table*}

%PACS
\begin{table*}[htbp]\centering
\caption{$acc_k$ \& $acc_u$ (\% Accuracy) on \textbf{PACS} Dataset .}\label{tab: PACS-uk}
\scriptsize
%\tiny

\begin{tabularx}{\textwidth}{l||YY|YY|YY|YY|YY}\toprule

\multirow{2}{*}{Method} &\multicolumn{2}{c}{Art Painting} &\multicolumn{2}{c}{Cartoon} &\multicolumn{2}{c}{Sketch} &\multicolumn{2}{c}{Photo} &\multicolumn{2}{c}{Average} \\\cmidrule{2-11}
&$acc_k$ &$acc_u$ &$acc_k$ &$acc_u$ &$acc_k$ &$acc_u$  &$acc_k$ &$acc_u$  &$acc_k$ &$acc_u$  \\\midrule
OSDAP \cite{28} &54.17 &49.84 &41.36 &51.68 &38.84 &54.92 &28.09 &41.62 &40.62 &49.51 \\
OpenMax \cite{29} &42.87 &91.48 &15.27 &97.44 &13.16 &96.61 &11.96 &90.22 &20.82 &93.94 \\
ERM \cite{27} &68.80 &24.57 &59.46 &33.08 &43.34 &20.27 &37.54 &30.03 &52.29 &26.99 \\
ERM+CM \cite{12} &68.66 &44.56 &62.25 &43.18 &41.01 &33.16 &39.91 &54.21 &52.96 &44.53 \\
ADA \cite{volpi2018generalizing} &70.95 &28.80 &62.08 &33.83 &43.18 &22.41 &40.65 &38.77 &54.22 &30.93 \\
ADA+CM \cite{12} &72.93 &40.12 &64.39 &49.06 &44.98 &40.85 &43.27 &52.53 &56.40 &45.64 \\
MEADA \cite{26} &70.90 &28.65 &62.09 &33.55 &43.42 &22.90 &39.78 &40.31 &54.05 &31.35 \\
MEADA+CM \cite{12} &70.45 &33.36 &63.76 &53.74 &40.25 &48.79 &42.89 &50.57 &54.34 &46.61 \\
\cmidrule{1-11}
\rowcolor[HTML]{d1ffff}\textsc{SODG-NET} &49.10 &69.17 &49.77 &65.65 &48.19 &74.09 &32.02 &71.68 &44.77 &70.15 \\
\rowcolor[HTML]{d1ffff}\textsc{SODG-NET} $-\mathcal{L}_{disc}$ &46.22 &72.15 &52.79 &59.45 &45.30 &77.95 &37.34 &51.93 &45.41 &65.37 \\
\rowcolor[HTML]{d1ffff}\textsc{SODG-NET} $-\mathcal{L}_{disc} - \mathcal{L}_{sm}$ &49.70 &64.02 &47.53 &55.49 &43.66 &76.02 &31.89 &53.86 &43.19 &62.35 \\
\bottomrule

\end{tabularx}
\end{table*}

\begin{table*}[!htp]\centering
\caption{Comparison of $\mathcal{F}_{ss}$ with MixStyle on \textbf{PACS} dataset}\label{tab: PACS-MixStyle}
\scriptsize
\begin{tabularx}{\linewidth}{Y||YY|YY|YY|YY|YY}\toprule
\multirow{2}{*}{Metric} &\multicolumn{2}{c}{Art Painting} &\multicolumn{2}{c}{Cartoon} &\multicolumn{2}{c}{Sketch} &\multicolumn{2}{c}{Photo} &\multicolumn{2}{c}{Average} \\\cmidrule{2-11}
&$\mathcal{F}_{ss}$ &MixStyle &$\mathcal{F}_{ss}$ &MixStyle &$\mathcal{F}_{ss}$ &MixStyle &$\mathcal{F}_{ss}$ &MixStyle &$\mathcal{F}_{ss}$ &MixStyle \\\midrule
$acc_k$ &\textbf{49.10} &44.44 &49.77 &\textbf{56.41} &\textbf{48.19} &45.36 &32.02 &\textbf{34.98} &44.77 &\textbf{45.30} \\
$acc_u$ &\textbf{69.17} &67.78 &\textbf{65.65} &46.65 &\textbf{74.09} &67.17 &\textbf{71.68} &46.04 &\textbf{70.15} &56.91 \\
$acc$ &\textbf{57.02} &53.61 &\textbf{56.01} &52.57 &\textbf{58.36} &53.93 &\textbf{46.27} &39.32 &\textbf{54.41} &49.86 \\
$hs$ &\textbf{57.44} &53.69 &\textbf{56.62} &51.07 &\textbf{58.40} &54.16 &\textbf{43.60} &39.75 &\textbf{54.02} &49.67 \\
\bottomrule
\end{tabularx}
\end{table*}

\section{Ablation Studies}

\noindent \textbf{Comparison of $\mathcal{F}_{ss}$ with MixStyle} \cite{zhou2021domain}: We conducted the experiments by replacing the style synthesis block $\mathcal{F}_{ss}$ with MixStyle method. The detailed results are shown in table \ref{tab: Office31-MixStyle} and table \ref{tab: PACS-MixStyle} for \textit{Office31} and \textit{PACS} dataset respectively. On \textit{Office31} dataset, we beat the results with MixStyle by $4.63\%$ and $4.84\%$  while on the \textit{PACS} dataset, on an average, we are outperforming the MixStyle by $4.55\%$ and $4.35\%$ in terms of $acc$ and $hs$ respectively.

\begin{table}[!htp]\centering
\caption{Comparison of $\mathcal{F}_{ss}$ with MixStyle on \textbf{Office31} dataset}\label{tab: Office31-MixStyle}
\scriptsize
\begin{tabularx}{\linewidth}{Y||YY}\toprule
Metric &$\mathcal{F}_{ss}$ &MixStyle \\\midrule
$acc_k$ &\textbf{73.96} &67.14 \\
$acc_u$ &\textbf{83.91} &80.43 \\
$acc$ &\textbf{79.02} &74.39 \\
$hs$ &\textbf{78.62} &73.78 \\
\bottomrule
\end{tabularx}
\end{table}

\noindent \textbf{Effects of changes in noise parameters in $\mathcal{F}_{ss}$}: To see the effects of added Gaussian noise to $\mu_1, \sigma_1, \mu_2, \sigma_2$ before passing them into $\mathcal{F}_{ss}$, we experiment with different values of $\mu, \sigma$ for the Gaussian distribution ($\mathbb{N}(\mu, \sigma)$). Table \ref{tab: Noise} shows the results of experiments on \textit{Office31} dataset.

\begin{table}[!htp]\centering
\caption{Effects of changes in noise parameters in $\mathcal{F}_{ss}$}\label{tab: Noise}
\scriptsize
\begin{tabularx}{\linewidth}{Y||Y|Y|Y|Y}\toprule
Metric &$\mathbb{N}(0,1)$ &$\mathbb{N}(1,1)$ &$\mathbb{N}(0,2)$ &$\mathbb{N}(0,3)$ \\\midrule
$acc_k$ &\textbf{73.96} &60.66 &73.41 &72.3 \\
$acc_u$ &\textbf{83.91} &81.5 &74.53 &80.7 \\
$acc $&\textbf{79.02} &71.25 &73.98 &76.57 \\
$hs$ &\textbf{78.62} &69.56 &73.96 &76.27 \\
\bottomrule
\end{tabularx}
\end{table}

\begin{table*}[htbp]\centering
\caption{{Results (\% Accuracy) on \textbf{DomainNet} Dataset for different source domains.}}\label{tab: DomainNet}
\scriptsize
%\tiny
\vspace*{-3mm}
\begin{tabularx}{\textwidth}{l||YY|YY|YY|YY|YY}\toprule\multirow{2}{*}{Method} &\multicolumn{2}{c}{Clipart} &\multicolumn{2}{c}{Painting} &\multicolumn{2}{c}{Sketch} &\multicolumn{2}{c}{Real} &\multicolumn{2}{c}{Average} \\

\cmidrule{2-11}
&$acc$ &$hs$ &$acc$ &$hs$ &$acc$ &$hs$ &$acc$ &$hs$ &$acc$ &$hs$ \\\midrule
ERM \cite{27} &27.21 &19.65 &23.14 &16.27 &35.83 &28.69 &40.18 &38.20 &31.59 &25.70 \\
ADA \cite{volpi2018generalizing} &32.42 &30.60 &36.65 &31.52 &38.97 &\textbf{31.23} &41.26 &40.65 &37.32 &33.50 \\
\cmidrule{1-11}
\rowcolor[HTML]{d1ffff}\textsc{SODG-Net} &\textbf{47.10} &\textbf{34.21} &\textbf{49.97} &\textbf{34.64} &\textbf{47.22} &27.55 &\textbf{49.68} &\textbf{46.19} &\textbf{48.49} &\textbf{35.65} \\
\bottomrule
\end{tabularx}
\vspace*{-3mm}
\end{table*}

\section{Closed set domain generalization}
In this section we provide the results on closed set single source domain generalization, that is, when the training and testing data label space is same. These experiments were conducted on two datasets, \textit{Office31} and \textit{PACS}. In case of the \textit{Office31} dataset, we have conducted an ablation study with varying number of classes in the dataset and source domain as Amazon (see table \ref{tab: CSSDG-OH}). For the \textit{PACS} dataset, each one of the four domains were taken as source domain for training and rest were considered as target. The average performance on the \textit{PACS} dataset for each case is given in table \ref{tab: CSSDG - PACS}. We compare performance of our style synthesis block against two baselines, one being the ERM \cite{27} and another one from Wang \etal \cite{wang2021learning}. We observe that our method performs convincingly when compared with the above mentioned ones.

\begin{table}[!htp]\centering
\caption{Closed set domain generalization on \textbf{Office31} dataset with Amazon as source domain with varying number of classes}\label{tab: CSSDG-OH}
\scriptsize
\begin{tabularx}{\linewidth}{l||YYY|Y}\toprule
\multicolumn{5}{c}{Office-31} \\\cmidrule{1-5}
Number of Classes &Amazon &DSLR &Webcam &Overall \\\midrule
31 &79.49 &50.75 &49.28 &69.9 \\
25 &82.53 &51.92 &52.63 &72.8 \\
20 &89.25 &59.76 &59.12 &79.56 \\
15 &89.22 &76.92 &62.5 &82.42 \\
10 &94.74 &77.5 &72.13 &88.06 \\
\bottomrule
\end{tabularx}
\end{table}

\begin{table}[!htp]\centering
\caption{{Closed set single source domain generalization results on \textbf{PACS} dataset}}\label{tab: CSSDG - PACS}
\scriptsize
\begin{tabularx}{\linewidth}{l||Y|Y|Y|Y|Y}\toprule
\multirow{2}{*}{Method} &\multicolumn{5}{c}{Source Domain} \\\cmidrule{2-6}
&Art &Cartoon &Photo &Sketch &Average \\\midrule
ERM \cite{27} &44.72 &41.96 &38.17 &29.24 &38.52 \\
Wang \etal \cite{wang2021learning} &\textbf{61.28} &66.35 &45.17 & 39.51& 53.07\\
\midrule
\rowcolor[HTML]{d1ffff} Our (SSB) &60.30 &\textbf{69.41} &\textbf{46.23} &\textbf{42.79} &\textbf{54.68} \\
\bottomrule
\end{tabularx}
\end{table}

{\small
\bibliographystyle{ieee_fullname}
\bibliography{egbib}
}